\documentclass{article} %
\usepackage{colm2024_conference}

\usepackage{microtype}
\usepackage{hyperref}
\usepackage{url}
\usepackage{booktabs}
\definecolor{darkblue}{rgb}{0, 0, 0.5}
\hypersetup{colorlinks=true, citecolor=darkblue, linkcolor=darkblue, urlcolor=darkblue}

\usepackage{url}
\usepackage{hyperref}
\usepackage{amsmath}
\usepackage{booktabs}
\usepackage{xspace}
\usepackage{pifont}
\usepackage{threeparttable}
\usepackage{tikz}
\usepackage{graphicx} 
\usepackage{pgfplots}
\usepackage{multirow}
\RequirePackage{algorithm}
\RequirePackage{algorithmic}
\usepackage{enumitem}
\usepackage{wrapfig}
\usepackage{xcolor,colortbl}
\usepackage{amssymb}
\usepackage{subcaption}

\usepackage{amsmath,amsfonts,bm}

\def\Figref#1{Figure~\ref{#1}}

\def\Secref#1{Section~\ref{#1}}

\def\eqref#1{equation~\ref{#1}}
\def\Eqref#1{Equation~\ref{#1}}

\def\1{\bm{1}}

\def\vh{{\bm{h}}}

\def\vp{{\bm{p}}}

\def\vs{{\bm{s}}}

\DeclareMathAlphabet{\mathsfit}{\encodingdefault}{\sfdefault}{m}{sl}
\SetMathAlphabet{\mathsfit}{bold}{\encodingdefault}{\sfdefault}{bx}{n}

\def\gL{{\mathcal{L}}}

\newcommand{\softmax}{\mathrm{softmax}}

\DeclareMathOperator*{\argmin}{arg\,min}

\newcommand\pawsemoji{\raisebox{-2pt}{\includegraphics[width=0.9em]{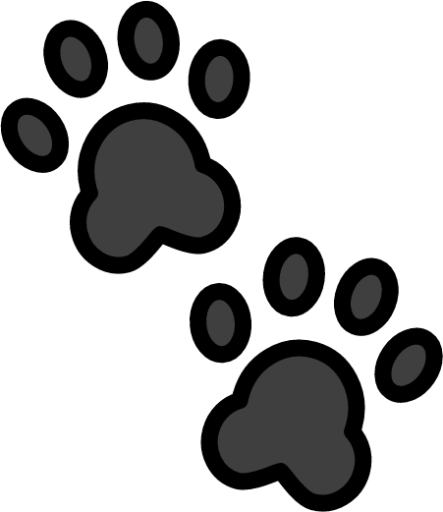}}}

\title{\pawsemoji{} Pack of LLMs: Model Fusion at Test-Time via  Perplexity Optimization}

\colmfinalcopy
\author{Costas Mavromatis \\
University of Minnesota\\
\texttt{mavro016@umn.edu} \\
\And
Petros Karypis \\
University of California San Diego\\
\texttt{pkarypis@ucsd.edu} \\
\AND
George Karypis \\
University of Minnesota\\
\texttt{karypis@umn.edu} \\
}

\newcommand{\packllm}{PackLLM\xspace}
\newcommand{\packs}{PackLLM$_\text{sim}$\xspace}
\newcommand{\packo}{PackLLM$_\text{opt}$\xspace}

\begin{document}

\maketitle

\begin{abstract}

Fusing knowledge from multiple Large Language Models (LLMs) can combine their diverse strengths to achieve improved performance on a given task. 
However, current fusion approaches either rely on learning-based fusers that do not generalize to new LLMs, or do not take into account how well each LLM understands the input. In this work, we study LLM fusion at \emph{test-time}, which enables leveraging knowledge from arbitrary user-specified LLMs during inference.  We introduce \emph{Pack of LLMs} (\packllm), an effective method for test-time fusion that leverages each LLM's expertise, given an input prompt. \packllm performs model fusion by solving an optimization problem for determining each LLM's importance, so that perplexity over the input prompt is minimized. 
First, our simple \packs variant validates that perplexity is a good indicator for measuring each LLM's expertise. 
Second, our \packo variant approximately solves the perplexity minimization problem via a greedy algorithm. The derived importance weights are used to combine the LLMs during inference.
We conduct experiments with over 100 total LLMs on a diverse set of tasks. Experimental results show that (i) perplexity is a reliable measure for LLM fusion, (ii) \packllm outperforms  test-time fusion baselines by 1.89\% accuracy points, and (iii) \packllm can leverage new LLMs to improve performance over learning-based fusion approaches by 3.92--11.94\% accuracy points. Our code is provided at \url{https://github.com/cmavro/PackLLM}.
\end{abstract}

\begin{figure}[h]
    \centering
    \includegraphics[width=.8\linewidth]{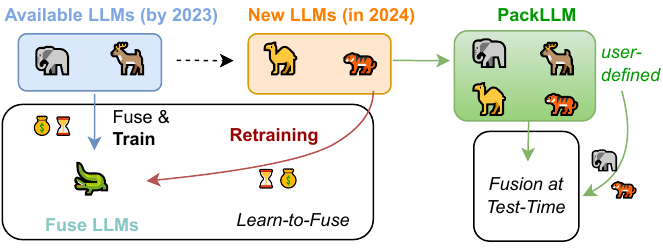}
    \caption{\packllm performs LLM fusion at \emph{test-time}. \packllm (i) does not require any training of fusion models, (ii) can leverage new (more powerful) LLMs when released, and (iii) allows users to select their preferred LLMs at test-time.}
    \label{fig:intro}
\end{figure}

\section{Introduction}

Large Language Models (LLMs)~\citep{brown2020fewshot,touvron2023llama} have achieved state-of-the-art performance on a variety of natural language processing (NLP) tasks and are beginning to be applied as Foundation Models~\citep{bommasani2021opportunities} for a gamut of applications.
As a result, we observe a sharp increase of newly released LLMs, that are pretrained on diverse corpora or fine-tuned to solve specific tasks. Moreover, it has become a strategic interest that corporations train and release their own LLMs, amplifying the number of available LLMs.  LLMs usually have architectural differences as well as they use different pretraining data, and as a result, their knowledge expertise is diversified. 
Therefore, an emergent question is ``\textit{How can we effectively combine knowledge of the available LLMs to improve downstream performance?}''.

Fusing knowledge from different models is a long-standing problem in machine learning~\citep{ho1995random,schapire2003boosting}. Regarding LLMs, current approach train dedicated fusion modules, such as neural networks~\citep{wang2023foe} or additional LLMs~\citep{jiang2023llm-blender,wan2024fusellm, ding2024mastering}, that learn to combine LLMs from a seed set. However, these approaches are not \emph{modular}, meaning they need to undergo expensive and time-consuming retraining to encompass newly released LLMs or remove specific LLMs, e.g.,  due to licensing issues. 

We present \emph{Pack of LLMs} (\packllm), a \emph{test-time} fusion method. As shown in \Figref{fig:intro}, \packllm does not require  any training of fusion modules, while it can combine arbitrary user-specified LLMs during inference. In order to fuse knowledge from the seed LLMs at test-time, \packllm performs a weighted ensemble of the output logits by posing an optimization problem at test-time. \packllm minimizes  the perplexity over the input prompt, so that  the fused LLM understands the task better. First, our simple \packs variant validates that perplexity is a good indicator for measuring each LLM's expertise. 
Second, our \packo variant approximately solves the perplexity minimization problem via a greedy algorithm. The derived importance weights are used for combining the LLMs during inference.

We conduct experiments with over 100 total LLMs and evaluate performance on language modeling tasks (\Secref{sec:exp-lang}) as well as on downstream tasks (\Secref{sec:exp-down}). Experiments show that \packllm's perplexity-based framework is a reliable indicator for determining fusion, outperforming approaches such as cBTM~\citep{gururangan2023cbtm} and DExperts~\citep{liu2021dexperts} that do not take into account how well the LLMs understand the input, while its overall performance scales better with respect to these baselines as we increase the number of expert LLMs. In addition, \packllm achieves a significant improvement of 1.72--1.89\% accuracy points (averaged over 25 tasks) over existing test-time fusion approaches, such as top expert selection and uniform ensemble. Furthermore, by employing newly released LLMs, \packllm outperforms competing learning-based fusion approaches, such as FuseLLM~\citep{wan2024fusellm} and FoE~\citep{wang2023foe}, by 3.92--11.94\% accuracy points, using as little as 3.92$\times$ fewer parameters. Our contributions are summarized below:
\begin{itemize}
    \item \textbf{Problem Formulation}: We study fusion of LLMs at test-time as a weighted ensemble and pose a perplexity minimization problem for determining the LLM importance weights.  
    \item \textbf{Algorithm}: We present \packo that solves the  perplexity minimization problem via a greedy algorithm. We also present \packs,  a simple, but effective, perplexity-based ensemble.
    \item \textbf{Effectiveness}: We evaluate \packllm on a diverse set of tasks showing that (i)  perplexity is a reliable indicator for model importance, (ii) \packllm outperforms  test-time fusion baselines by 1.72--1.89\% accuracy points, (iii) \packllm can leverage newly released LLMs and outperform  competing learning-based fusion approaches by 3.92--11.94\% accuracy points.
\end{itemize}

\section{Related Work}

 Model fusion is a long-standing problem in machine learning~\citep{ho1995random,sollich1995learning,kuncheva2003measures,schapire2003boosting}. Regarding fusion of LLMs, \emph{weight merging}~\citep{yadav2023ties} and \emph{mixture-of-experts}~\citep{komatsuzaki2022sparse} are techniques that fuse knowledge stored in LLMs by combining their model parameters.
However, such approaches require that the given LLMs share the same architecture, which is limited with the current availability of  powerful LLMs of varying sizes. An alternative technique is \emph{majority voting}~\citep{li2024agents}, but can only be applied to certain downstream tasks, such as classification tasks.

\begin{figure*}[tb]
    \centering
    \includegraphics[width=0.85\linewidth]{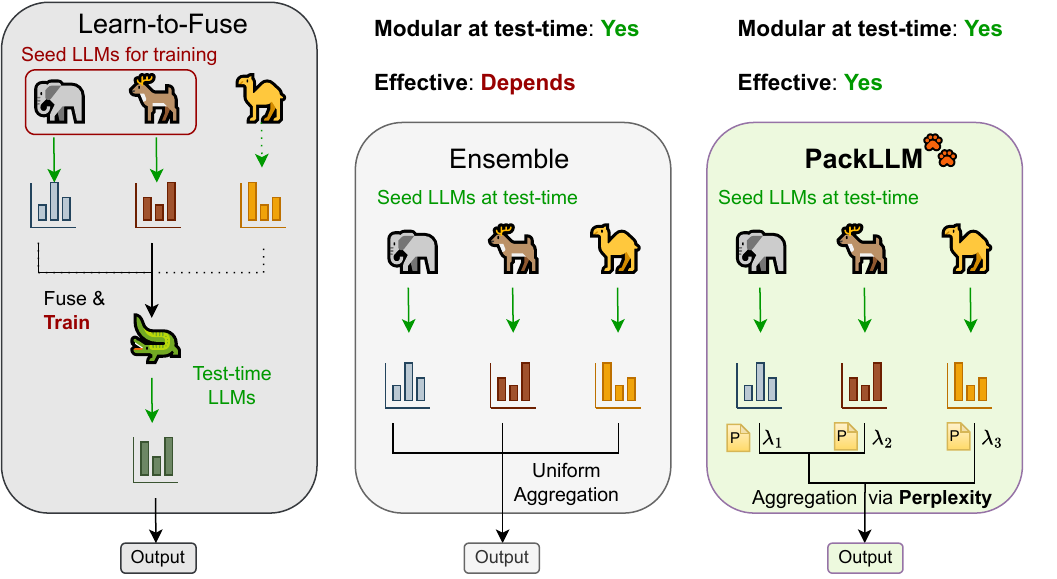}
    \caption{Overview of LLM fusion paradigms. \textit{Left}: Learning-based fusion approaches are not modular and cannot encompass new LLMs at test-time. \textit{Middle}: Uniform ensemble may degrade performance, given weak seed LLMs. \textit{\textbf{Right}}: Our \textbf{\packllm} approach determines the weights of the seed LLMs at test-time via perplexity to achieve effective performance.}
    \label{fig:fusion}
\end{figure*}

Recent successful fusion approaches often follow a \emph{learn-to-fuse} framework (see \Figref{fig:fusion}, left), where specialized learnable modules~\citep{wang2023foe,sukhbaatar2024branch} or even LLMs~\citep{jiang2023llm-blender,wan2024fusellm,ding2024mastering}, are trained to combine knowledge from a set of seed LLMs. However, these approaches are \emph{not modular}, meaning they cannot generalize to arbitrary user-specified LLMs and require additional retraining when models need to be added in or removed from the seed set. Approaches, such as ~\citep{feng2024knowledgecard}, fuse the generations of different LLMs, but require a strong (large-scale) LLM as a fuser, which is computationally expensive.

On the other hand, \emph{model ensemble} approaches~\citep{liu2021dexperts,gururangan2023cbtm,liu2024proxy,li2024purifying} perform fusion at the output-level, i.e., output token logits or probabilities. However,  effectively combining the LLMs during test-time, i.e., accurately determining each LLM's importance, has not been extensively studied as these approaches do not take into account how well each LLM understands the input. For instance, uniform ensemble may degrade performance if the set includes weak LLMs (\Figref{fig:fusion}, middle).

\section{Pack of LLMs (\packllm) \pawsemoji{}} \label{sec:packllm}

In this work, we study the following problem: 
\begin{center}
\fbox{\parbox{5.5in}{\centering
Given a set of user-specified LLMs $\{M_1, \dots, M_K\}$ at test-time, how can we effectively combine the knowledge of each LLMs to achieve improved performance?}}
\end{center}

We present Pack of LLMs (\packllm), a weighted ensemble of LLMs that fuses knowledge of arbitrary LLMs at test-time. As shown in \Figref{fig:fusion} (right), given a set of seed LLMs $\{M_1, \dots, M_K\}$ at test-time, \packllm fuses their knowledge with corresponding importance weights $\{\lambda_1, \dots, \lambda_K\}$. The importance weights $\lambda_k$ are calculated based on how well each LLM understands the prompt $P$.

In more detail, at each step $t$, each model $M_k$ is conditioned on the prompt $x_{< t}$, to obtain the logit score $s_{k}(x_t | x_{< t})$ for the next token $x_t$. Logit scores $\vs_{k}$ are unnormalized scores over the LLM's vocabulary, before  applying the $\softmax(\cdot)$  operation over all vocabulary tokens
which transforms them into probabilities $\vp_{M_k}$. Here, we assume the models share the same vocabulary, but we provide a generalized version in \Secref{sec:tokenize}.
\packllm combines output logits to obtain the final probabilities as
\begin{equation}
    p(x_t | x_{< t}) = \softmax \big(\sum_{k=0}^K \textcolor{teal}{\underbrace{\textcolor{black}{\lambda_k}}_{unknown}} s_{k}(x_t | x_{< t}) \big),
    \label{eq:fusion}
\end{equation}
where we assume that the weights $\lambda_k \in [0,1]$ are normalized, i.e., $\sum_{k=0}^K \lambda_k = 1$.
In what follows, we answer the question ``\textit{How can we accurately determine weights $\{\lambda_1, \dots, \lambda_K\}$ at test-time?}''

\subsection{Prompt Perplexity Minimization}
\packllm relies on the notion that perplexity is a suitable measure for a LLM's knowledge on a certain input (where the lower the better). Perplexity is a metric that aligns with the causal language modeling objective of LLMs, enabling to measure whether a test input falls into the LLM's expertise and how the test input relates to the LLM's pretraining data~\citep{marion2023less,gonen2022demystifying}.

Given a tokenized input prompt or sequence $P = (x_0, x_1, \dots, x_t)$, perplexity (PPL) is defined as the exponentiated average negative log-likelihood of $P$ as
\begin{equation}
    \text{PPL}_k(P) = \exp  \big\{ -\frac{1}{t} \sum_i^t \log p_{k}(x_i| x_{<i} ) \big\},
    \label{eq:ppl}
\end{equation}
where $\log p_k(x_i| x_{<t} )$ is the log-likelihood of the $i$-th token conditioned on the preceding tokens $x_{<t}$ according to model $M_k$. Note that PPL depends solely on the models and the input prompt and can be evaluated at test-time. 

Leveraging perplexity as a measurement for a LLM's expertise on the test prompt, we formulate the assignments of importance weights as an optimization problem that does not require any training or annotated data and can be solved at test-time. 
Formally, our optimization framework is given as
\begin{equation}
    \begin{aligned}
    \lambda^\ast_1, \dots,  \lambda^\ast_K = \argmin_{\lambda_1, \dots,  \lambda_K} & \Big( -\frac{1}{t} \sum_i^t \log \; \underbrace{\softmax \big( \sum_{k=0}^K \lambda_k s_{k}(x_i| x_{<i} )}_{p(x_i| x_{<i} )} \big) \Big), \\
    \text{subject to} & \;  \lambda \in [0,1] \text{ and }  \; \sum_k^K \lambda_k = 1,
    \end{aligned}
    \label{eq:optimize}
\end{equation}
where $\{\lambda^\ast_1, \dots,  \lambda^\ast_K\}$ are the optimal importance weights and $\vs_k$ are the output logits of each LLM. 
During inference, the computed weights $\{\lambda^\ast_1, \dots,  \lambda^\ast_K\}$ are used in \Eqref{eq:fusion}.

\subsection{\packs: Simple Perplexity-Based Weighting} \label{sec:sppl}

An important question is whether perplexity is a reliable metric 
for determining fusion.
As the first step, we present \packs, a simple perplexity-based approach for determining the importance weights $\{\lambda_1, \dots, \lambda_K\}$. Instead of solving the optimizing problem presented in \Eqref{eq:optimize}, \packs computes the weights $\lambda_k$ directly by using the perplexity scores over the prompt via
\begin{equation}
    \lambda_k = \softmax_k \Big( - \log \text{PPL}_k(P) / \tau \Big),
    \label{eq:packs}
\end{equation}
where $\tau$ is the temperature hyperparameter with default values $\tau \in \{0.1, 1\}$. Weights $\{ \lambda_1, \cdots, \lambda_k\}$ are inversely proportional to the models' normalized perplexity scores, i.e., lower perplexity yields a higher weight. 
\packs serves as a validation of whether the models' perplexity scores provide useful information for determining fusion.

\subsection{\packo: Greedy Perplexity Optimization} \label{sec:ppl}
Unlike the simple heuristic that \packs uses (\Eqref{eq:packs}), \packo approximately solves the perplexity minimization problem presented in \Eqref{eq:optimize}. 
The problem in \Eqref{eq:optimize} is overdetermined, where the number of equations is $t$ ($t$ prompt tokens) and the number of variables is $K$ ($K$ seed LLMs), having many solutions. Nevertheless, it can be solved at test-time via grid search~\citep{lavalle2004grid} by evaluating different values $\lambda_k \in [0, 1]$ with the constraint $\sum_k \lambda_k = 1$. However, exhaustive grid search is time-consuming during inference due to its combinatorial complexity on the number of LLMs.

We propose an efficient greedy algorithm, which ensembles the LLMs sequentially. The sequential nature of the algorithm reduces the optimization problem to searching between linear combinations of two models rather than the full $K$ models.

First, we compute the perplexity $\text{PPL}_k(P)$ of each LLM $k$ via \Eqref{eq:ppl}
and sort the seed LLMs $\{M_1, \dots, M_K\}$ based on $\text{PPL}_k(P)$ as
\begin{equation}
    [ M^\ast_1, \dots, M^\ast_K] = argsort \big( \text{PPL}_1(P), \dots, \text{PPL}_K(P) \big),
    \label{eq:sort}
\end{equation}
where $[ M^\ast_1, \dots, M^\ast_K]$ are ordered by the lowest to highest perplexity scores. By using the ordered set, we can omit using irrelevant LLMs via early stopping of the sequential ensemble. 

At the first step of the greedy optimization, we determine the relative weights between the top-1 and the top-2 LLMs, $ M^\ast_1$ and $M^\ast_2$. We solve 
\begin{equation}
    \lambda^{(1)} = \argmin_{\lambda} \Big( -\frac{1}{t} \sum_i^t \log \softmax \big( \lambda s^{(1)}(x_i| x_{<i} ) + (1-\lambda) s^{(2)}(x_i| x_{<i} ) \big) \Big),
    \label{eq:greedy1}
\end{equation}
where $\vs^{(1)}$ and $\vs^{(2)}$ are the output logits by $ M^\ast_1$ and $M^\ast_2$, respectively, and $\lambda \in [0, 1]$.
The optimization in \Eqref{eq:greedy1} is solved via grid search (greedy grid search), where different values of  $\lambda \in [0, 1]$ are evaluated (the default step is $0.05$). 
The value $\lambda^{(1)}$ that results to the lowest perplexity is used to update
\begin{equation}
    \vs^{(2)} = \lambda^{(1)} \vs^{(1)} + (1-\lambda^{(1)}) \vs^{(2)}
\end{equation}
for the next step. 

The same procedure is repeated to iterate through all seed LLMs $[ M^\ast_1, \dots, M^\ast_K]$. We can perform early stopping when we find $\lambda^{(k)} = 1$, i.e., the effect of the current LLM is zero.
Our overall greedy optimization is summarized in Appendix~\ref{app:alg}.

\subsection{Tokenizer Selection \& Alignment} \label{sec:tokenize}

\Eqref{eq:fusion} assumes that the logit vectors $\vs_k$ can be added together, which requires that the seed LLMs share the same vocabulary. However, this is not always the case, as different LLMs are trained based on different tokenizers. To address this, we employ a tokenizer selection and alignment strategy for combining LLMs with different tokenizers.

First, we determine the reference tokenizer as the tokenizer of the top-1 LLM (\Eqref{eq:sort}). 
Then, we follow the Minimum Edit Distance (MinED) approach~\citep{fu2023specializing,wan2024fusellm}, where each token of the reference tokenizer is mapped to its textually closest token of another tokenizer, e.g., ``get'' to ``gets''. We tokenize the sequence with the reference tokenizer and use the MinED token mappings to obtain a valid input for another LLM. With the new input, we obtain the output logits of the LLM over its vocabulary. Then, we map the output logits back to the reference tokenizer's vocabulary via MinED in order to perform logit fusion via \Eqref{eq:fusion}. MinED mappings between tokenizers are precomputed before inference. We note that
aligning different tokenizers is an open problem in NLP, and  more effective alignment techniques may further improve \packllm.

\section{Language Modeling Experiments} \label{sec:exp-lang}

The first research question that we  answer is: \\
\textbf{RQ1}. \textit{Is  perplexity a good measure for determining the LLM importance weights at test-time?}

We compare with expertise-based model ensemble approaches cBTM\citep{gururangan2023cbtm} and DExperts~\citep{liu2021dexperts}, which, unlike \packllm, do not rely on perplexity. 

\textbf{cBTM} performs model fusion at the output probabilities, where the weight of each expert is determined based on the tf-idf similarity between the given prompt and the expert's pretraining data. In particular, cBTM computes weights $\{\lambda_1, \dots, \lambda_K\}$ for a test prompt $P$ for test-time fusion as
\begin{equation}
    \lambda_k \propto \softmax \big( -d(\vh_P , \vh_{k} ) \big) \; \text{and } \;  p(X_t | x_{< t}) = \sum_k \lambda_k p_k(X_t | x_{< t}),
    \label{eq:cbtm}
\end{equation}
where $d(\cdot, \cdot)$ is a Euclidean distance,  $\vh_P$ is the tf-idf embedding of the prompt, and $\vh_{k}$ is a representation of the pretraining data of model $k$. Note that \Eqref{eq:cbtm} does not use perplexity, but relies on data embedding similarity. 

\textbf{DExperts} and its follow-up works~\citep{liu2024proxy,li2024purifying} employ an additional base model $M$ and an anti-expert model $M^{-}$ to fuse knowledge from an expert model. In particular, the fusion is performed as 
\begin{equation}
    \vp = \softmax \big( \vs_M + \lambda  [\vs_{M^\ast_1} - \vs_{M^{-}}]\big),
\end{equation}
where $M^\ast_1$ is the top-1 expert from the seed LLMs, and $\lambda$ is a hyperparameter (default value is $\lambda=1$). DExperts uses fixed LLM weights. 

\subsection{Experimental Setup}

 We follow the cBTM experimental setting, where C4~\citep{roberts2019exploring} and S2ORC~\citep{lo2019s2orc} datasets are clustered into $K \in \{1, 2,4,8,16\}$ clusters, and $K$ OPT-1.3B models~\citep{zhang2022opt} are trained on each cluster separately. Therefore, each OPT model is an expert on a specific semantic cluster of the training documents. The clusters are computed based on the tf-idf embeddings of the train data and the embedding of the cluster centroid is used as $\vh_k$ in \Eqref{eq:cbtm} of cBTM. Overall, we experiment with a total of \underline{65 LLMs}. To compare  with DExperts, we employ the OPT-6.7B, OPT-13B and OPT-30B models as additional base models $M$. Following DExperts, the anti-expert is the original OPT-1.3B model. 
 For each dataset, we report language modeling perplexity on 200 randomly-sampled held out documents. For each test document, the prompt $P$ consists of the 32 first tokens or the first 20\% of the sequence's  tokens, whichever is shorter;  perplexity is evaluated on the rest of tokens.

\begin{figure*}[t!]
  \centering
  \begin{subfigure}[t]{0.48\linewidth}
      \resizebox{\linewidth}{!}{\definecolor{col1}{rgb}{0.60, 0.31, 0.64}
\definecolor{col2}{rgb}{0.30, 0.69, 0.29}
\definecolor{col3}{rgb}{0.22, 0.49, 0.72}
\definecolor{col3b}{rgb}{0.25, 0.55, 0.70}
\definecolor{col4}{rgb}{0.89, 0.10, 0.11}
\definecolor{col5}{rgb}{1, 1, 0.8}

\begin{tikzpicture}
\tikzstyle{every node}=[font=\huge]

\begin{axis}[legend style={at={(.9,0.9),anchor=north east}},
             legend style={legend pos=outer north east,},  title={C4},ylabel={Perplexity $\downarrow$}, xlabel={\#LLMs ($K$)}, every axis plot/.append style={ultra thick}, ymin=11, ymax=14, compat=1.5, xtick={2, 4, 8, 16}
]

\addplot[mark=*,orange, error bars/.cd, y dir=both, y explicit] coordinates {
    (2, 13.35)
    (4, 12.64) 
    (8, 12.31)
    (16, 12.23)
};

\addplot[mark=diamond*,col3, error bars/.cd, y dir=both, y explicit] 
coordinates {
    (2, 12.82)
    (4, 12.22) 
    (8, 11.68)
    (16, 11.41)
};

\addplot[mark=square*,col2, error bars/.cd, y dir=both, y explicit] 
coordinates {
    (2, 12.82)
    (4, 12.13) 
    (8, 11.67)
    (16, 11.45)
};

\end{axis}

\begin{axis}[legend columns=-1,legend style={at={(0.7,1.35),anchor=north east,}},
              title={S2ORC}, xlabel={\#LLMs ($K$)}, every axis plot/.append style={ultra thick}, ymin=11.5, ymax=13, compat=1.5, xtick={2, 4, 8, 16}, xshift=8.5cm
]

\addplot[mark=*,orange, error bars/.cd, y dir=both, y explicit] coordinates {
    (2, 12.79)
    (4, 12.42) 
    (8, 12.23)
    (16, 12.52)

};

\addlegendentry{cBTM \; }

\addplot[mark=diamond*,col3b, error bars/.cd, y dir=both, y explicit] 
coordinates {
    (2, 12.82)
    (4, 12.32) 
    (8, 12.01)
    (16, 11.83)
};
\addlegendentry{\textbf{\packs} \; }

\addplot[mark=square*,col2, error bars/.cd, y dir=both, y explicit] 
coordinates {
    (2, 12.82)
    (4, 12.38) 
    (8, 12.14)
    (16, 12.10)
};
\addlegendentry{\textbf{\packo} }

\end{axis}

\end{tikzpicture}}
    \caption{$K$ expert models.}
     \label{fig:cbtm}
  \end{subfigure}
   \begin{subfigure}[t]{0.51\linewidth}
      \resizebox{\linewidth}{!}{\definecolor{col1}{rgb}{0.60, 0.31, 0.64}
\definecolor{col2}{rgb}{0.30, 0.69, 0.29}
\definecolor{col3}{rgb}{0.22, 0.49, 0.72}
\definecolor{col4}{rgb}{0.89, 0.10, 0.11}
\definecolor{col5}{rgb}{1, 1, 0.8}

\begin{tikzpicture}
\tikzstyle{every node}=[font=\huge]

\begin{axis}[legend style={at={(.9,0.9),anchor=north east}},
             legend style={legend pos=outer north east,},  title={C4},ylabel={Perplexity $\downarrow$}, xlabel={\#LLMs ($K$)}, every axis plot/.append style={ultra thick}, ymin=10.8, ymax=12.6, compat=1.5, xtick={2, 3, 5, 9}
]

\addplot[mark=*,orange, error bars/.cd, y dir=both, y explicit] coordinates {
    (2, 12.31)
    (3, 11.92)
    (5, 12.11) 
    (9, 12.04)

};

\addplot[mark=triangle*,col1, error bars/.cd, y dir=both, y explicit] coordinates {
    (2, 12.22)
    (3, 12.20)
    (5, 12.12) 
    (9, 12.01)

};

\addplot[mark=diamond*,col3, error bars/.cd, y dir=both, y explicit] 
coordinates {
    (2, 12.16)
    (3, 11.97) 
    (5, 11.54)
    (9, 11.17)

};

\addplot[mark=square*,col2, error bars/.cd, y dir=both, y explicit] 
coordinates {
    (2, 11.93)
    (3, 11.75)
    (5, 11.37) 
    (9, 11.03)

};

\end{axis}

\begin{axis}[legend columns=-1,legend style={at={(1,1.35),anchor=north east,}},
              title={S2ORC},, xlabel={\#LLMs ($K$)}, every axis plot/.append style={ultra thick}, ymin=11, ymax=12.2, compat=1.5, xtick={2, 3, 5, 9}, xshift=8.5cm
]

\addplot[mark=*,orange, error bars/.cd, y dir=both, y explicit] coordinates {
    (2, 11.83)
    (3, 11.23)
    (5, 11.48) 
    (9, 11.64)

};

\addlegendentry{cBTM \;}

\addplot[mark=triangle*,col1, error bars/.cd, y dir=both, y explicit] coordinates {
    (2, 11.59)
    (3, 11.52)
    (5, 11.45) 
    (9, 11.40)

};

\addlegendentry{DExperts \; }

\addplot[mark=diamond*,col3, error bars/.cd, y dir=both, y explicit] 
coordinates {
    (2, 12.06)
    (3, 11.89)
    (5, 11.71)
    (9, 11.52)

};
\addlegendentry{\textbf{\packs} \; }

\addplot[mark=square*,col2, error bars/.cd, y dir=both, y explicit] 
coordinates {
    (2, 11.56)
    (3, 11.39)
    (5, 11.18)
    (9, 11.05)
};
\addlegendentry{\textbf{\packo} }

\end{axis}

\end{tikzpicture}}
    \caption{$K$ expert models combined with one large model.}
     \label{fig:dexperts}
  \end{subfigure}
  \caption{Perplexity result comparison (the lower, the better) on C4 and S2ORC datasets.}
\end{figure*}
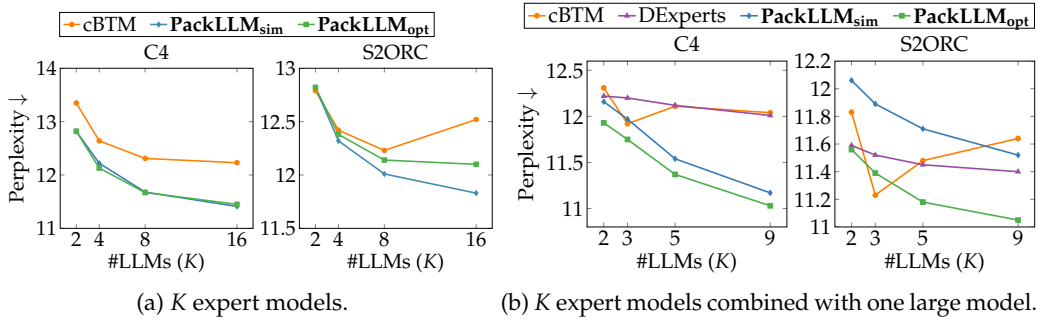

\subsection{Results}

\Figref{fig:cbtm} shows perplexity results when we compare \packllm with the weighted ensemble approach of cBTM. \packllm outperforms cBTM under all values of  the number of expert LLMs $K$. The performance differences grow as the number of LLMs increases  and the largest improvement is observed when we have $K=16$ experts. \packllm takes into account the perplexity of the underlying experts to estimate their expertise, while cBTM relies on off-the-shelf similarity embeddings and may not successfully capture the LLMs' knowledge. \packs and \packo perform similarly on the C4 dataset, while \packs scales better on S2ORC dataset, which has shorter prompts. The results show that our perplexity-guided fusion is effective and scales better as we increase the number of expert models. 

\Figref{fig:dexperts} shows averaged perplexity results when we add additional large models (OPT-6.7B OPT-13B, OPT-30B) to the set of seed expert LLMs (C4/S2ORC-tuned OPT1.3B models), evaluating whether the fusion approaches can effectively incorporate larger general-purpose LLMs. \packllm outperforms cBTM and DExperts, and the improvement becomes larger as we increase the number of expert LLMs (value $K$). Moreover, \packo outperforms \packs in both C4 and S2ORC; its benefit lies on its perplexity optimization, which effectively combines knowledge from expert models with larger  models to boost performance during inference. The result show that \packllm is effective with LLMs of varying sizes and expertise.

\section{Downstream Tasks Experiments} \label{sec:exp-down}
The following research questions that we answer are: \\
\textbf{RQ2}. \textit{How does \packllm compare against other test-time fusion approaches in downstream tasks?} \\
\textbf{RQ3}. \textit{What is the benefit of \packllm over learning-based fusion approaches?} 

\subsection{Experimental Setup}

\textbf{Datasets}. We experiment with a diverse set of tasks. We employ the \textbf{MMLU-STEM}~\citep{hendrycks2021mmlu} dataset to evaluate performance in knowledge intensive tasks, such college STEM exams.  We employ ARC~\citep{clark2018arc}, BoolQ~\citep{clark2019boolq}, HellaSwag~\citep{zellers2019hellaswag}, and OpenBookQA~\citep{mihaylov2018openbook} datasets to evaluate performance on \textbf{Commonsense} reasoning. Furthermore, we employ MMLU-Health, PubMedQA~\citep{jin2019pubmedqa}, USMLE~\citep{jin2021usmle} datasets to evaluate performance on domain-specific knowledge (\textbf{Medicine}). In addition, we evaluate on miscellaneous (\textbf{Misc.}) classification tasks~\citep{mavromatis2023adaicl}, such as topic classification, hatespeech detection, and sentiment analysis. Further details can be found in Appendix~\ref{app:data}. We report accuracy on question answering and classification as the evaluation metric. 

\textbf{Seed LLMs}. For MMLU-STEM and Commonsense tasks, we employ widely used LLMs, such as Mistral-7B~\citep{jiang2023mistral} and LLaMa2-7B~\citep{touvron2023llama}.  For Medicine tasks, we employ specialized LLMs for biomedical data, such as AdaptLLM~\citep{cheng2023adaptllm} and BioMistral~\citep{labrak2024biomistral}. For Misc. tasks, we use models with complementary expertise~\citep{gururangan2023cbtm}. Overall, we experiment with a total of  \underline{52 LLMs} (see Appendix~\ref{app:llms}).

\textbf{Baselines}. We compare with the following test-time and learning-based fusion approaches. 
(i) \textbf{Top1-PPL} selects the most relevant LLM for a test prompt $P$. We use prompt perplexity (\Eqref{eq:sort}) to determine the top-1 LLM. (ii) \textbf{Ensemble} performs uniform aggregation of the LLM's logits where every weight $\lambda_k$ has value  $\lambda_k = \frac{1}{K}$. (iii) \textbf{FuseLLM}~\citep{wan2024fusellm} uses the output logits of the seed LLMs to train a fusion LLM via knowledge distillation. However, it requires an external corpus  for causal language modeling pretraining and cannot encompass new LLMs at test-time. (iv) \textbf{FoE}~\citep{wang2023foe} trains a neural network classifier to select the most relevant expert from the seed LLMs for a test prompt. However, it requires annotated data, e.g., validation data for the downstream tasks, which limits its applicability to new tasks and LLMs. We also compare with (v) \textbf{cBTM} and (vi) \textbf{DExperts} when possible, as they cannot handle arbitrary seed LLMs. 

\newcommand{\mc}[2]{\multicolumn{#1}{c}{#2}}
\definecolor{Gray}{gray}{0.85}
\definecolor{LightCyan}{rgb}{0.9,0.95,1}
\definecolor{LightPink}{rgb}{0.97,0.92,0.92}
\definecolor{LightGreen}{rgb}{0.9,0.95,0.9}
\definecolor{LightOrange}{rgb}{0.99,0.95,0.75}
\newcolumntype{a}{>{\columncolor{LightPink}}c}
\newcolumntype{b}{>{\columncolor{LightCyan}}c}
\newcolumntype{d}{>{\columncolor{LightGreen}}c}
\newcolumntype{e}{>{\columncolor{LightOrange}}c}

\begin{table*}
\centering
\caption{Main Results. \packllm outperforms  other test-time fusion approaches in 25 tasks and with different seed LLMs. }
\label{tab:main}%
\resizebox{\columnwidth}{!}{
\begin{threeparttable}
    \begin{tabular}{laaabbbeedc}
        \toprule
        \multirow{2}{*}{}& \multicolumn{3}{a}{\textbf{MMLU STEM}} & \multicolumn{3}{b}{\textbf{Commonsense}} & \multicolumn{2}{e}{\textbf{Medicine}} & \multicolumn{1}{d}{\textbf{Misc.}} &
        \textbf{Average} \\
        & \multicolumn{3}{c}{\textit{5 tasks (5-shot)}} & \multicolumn{3}{c}{\textit{4 tasks (0-shot)}} & \multicolumn{2}{c}{\textit{10 tasks (0/5-shot)}} & \multicolumn{1}{c}{\textit{6 tasks (5-shot)}}  & \\
        \midrule
        \multirow{1}{*}{\emph{Test-Time Fusion}} &
        \multicolumn{9}{c}{\textit{\#Seed LLMs $K$ (\#Total Parameters)}} \\
        \rowcolor{white}
        &  2 (14B) & 4 (23.7B) & 6 (34.7B) & 2 (14B) & 4 (23.7B) & 6 (34.7B) & 3 (27B) & 4 (28B) & 12 (15.6B) & \\
        \cline{2-10}
        cBTM & -- & -- & -- & -- & -- & -- & -- & -- & 61.95 & --\\
        DExperts & -- & -- & -- & -- & -- & -- & 44.37 & -- & -- & --\\
        Ensemble & 41.29 & 	43.54 & 43.51 & 66.30 & 64.45 & 62.10 &  42.22 & 51.59 & 62.53 & 53.05\\
        Top1-PPL & 37.59 & 40.73 & 44.67 & 61.32 & 64.45 & 	65.31 &  46.83 & 50.11 & 63.95 & 52.77\\
        \textbf{\packs} & 42.16 & 44.22 & 44.48 & 63.84 & 69.23  &  68.50 &43.55 & 53.73& 63.20 & \underline{54.77}\\
        \textbf{\packo} & 42.30 & 45.02 & 45.60 & 61.81 & 67.67 & 67.58 & 47.21 & 53.88  & 63.38&\textbf{54.94}\\
        \bottomrule
    \end{tabular}%
    \begin{tablenotes}
        \item -- : cBTM and DExperts cannot handle arbitrary seed LLMs.
    \end{tablenotes}

    \end{threeparttable}
}

\end{table*}%

\subsection{Results}

Table~\ref{tab:main} compares \packllm with competing test-time approaches, such as top-1 expert selection based on perplexity (Top1-PPL) and uniform ensemble (Ensemble). \packllm outperforms these approaches by 1.72--1.89\% accuracy points, averaged over different  seed LLMs in 25 tasks. The improvement becomes larger when using a greater number of LLMs, e.g., in commonsense (+3.19\% accuracy points) and medicine  (+2.29\% accuracy points) tasks. Uniform ensemble may be negatively impacted when we increase the number of LLMs (not all LLMs are equally strong), while Top1-PPL improves with more LLMs. \packo outperforms \packs in all tasks, except for Commonsense, where the input prompts are shorter (0-shot input). Moreover, \packllm outperforms cBTM and DExperts in the cases they can compare against. The results show that \packllm is an effective fusion approach, outperforming competing test-time approaches by up to 1.89\% accuracy points, on average.

\begin{table}
\centering
\caption{Results on LLM fusion approaches. \packllm leverages recently released LLMs and outperforms learning-based approaches. }
\label{tab:fusellm}%
\resizebox{\columnwidth}{!}{
\begin{threeparttable}
    \begin{tabular}{l|cccc|c}
        \toprule
        
        \textit{Seed LLMs  $\rightarrow$} & \multicolumn{1}{a}{FuseLLM LLMs } & \multicolumn{1}{b}{FoE LLMs} & \multicolumn{1}{e}{\underline{Recent LLMs}}  & \multicolumn{1}{d}{\underline{Recent SLMs$^\ast$}}  &  Best Overall \\

        \textit{\; \; ($K$, \#Params)}& \multicolumn{1}{a}{(3, 21B)} & \multicolumn{1}{b}{(15, 93B)} & \multicolumn{1}{e}{(4, 23.7B)} & \multicolumn{1}{d}{(3, 6.5B)} & \\

        \textit{ \; \;release date}& \multicolumn{1}{a}{\textit{before 07/2023}} & \multicolumn{1}{b}{\textit{before 08/2023}} & \multicolumn{1}{e}{\textit{07-12/2023}} & \multicolumn{1}{d}{\textit{12/2023-02/2024}} \\
        
        \midrule
        & STEM / Com. & STEM / Com. & STEM / Com. & STEM / Com. & STEM / Com.\\
        \midrule
        \multicolumn{1}{l}{\textit{Learning-based fusion}} & \\
        FuseLLM-7B~\citep{wan2024fusellm} & 33.08  / 58.59 &  ($\circlearrowright$) / ($\circlearrowright$) &  ($\circlearrowright$) / ($\circlearrowright$) & ($\circlearrowright$) / ($\circlearrowright$) & \multicolumn{1}{a}{33.08 / 58.59}  \\
        
        FoE~\citep{wang2023foe} & \; \; ($\circlearrowright$) / ($\circlearrowright$, $\dagger$) & 40.65 /  ($\dagger$) \; \; & \; \; ($\circlearrowright$) / ($\circlearrowright$, $\dagger$) & \; \; ($\circlearrowright$) / ($\circlearrowright$, $\dagger$) & \multicolumn{1}{b}{40.65 / \;\; -- \; \;} \\
        \midrule
        \multicolumn{1}{l}{\textit{Test-time fusion}} & \\
        \textbf{\packllm}  & 31.21 / 54.88 & 40.26 / 67.05 & 45.02 / 69.23 & 37.00 / 68.25 & \multicolumn{1}{e}{\textbf{45.02 / 69.23}}  \\
    
        \bottomrule
    \end{tabular}%
    \begin{tablenotes}
    \item $\ast$ : We use the term SLMs (Small Language Models) to emphasize that their size is below 3B.
        \item $\circlearrowright$ : These  approaches require retraining with the new seed LLMs before test-time fusion.
        \item $\dagger$ : These approaches require additional annotated data for training.
    \end{tablenotes}

    \end{threeparttable}
}

\end{table}%

In Table~\ref{tab:fusellm}, we compare \packllm with learning-based fusion approaches. FuseLLM trains a fusion LLM that learns to combine knowledge from the seed LLMs over diverse pretraining data, while FoE trains a classifier to select the best expert LLM based on validation data. As a result, both FuseLLM and FoE have an advantage over our learning-free method when using the same seed LLMs, leading to 0.39--3.71\% more accuracy points. However, \packllm does not require any training or validation data and can quickly adapt as new LLMs are released. For example, \packllm improves over FuseLLM and FoE by 4.37--11.94\% accuracy points when using four recently released 7B LLMs. In addition, by using strong small LMs~\citep{li2023phi}, \packllm outperforms FuseLLM by 3.92--10.66\% accuracy points without incurring additional computational costs (equal number of total LLM parameters). Full results of this section can be found in Appendix~\ref{app:full-results}.

\section{Analysis}

In this section, we analyze \packs and \packo with respect to their derived LLM importance weights, their sensitivity on the input prompt, and their time complexity.

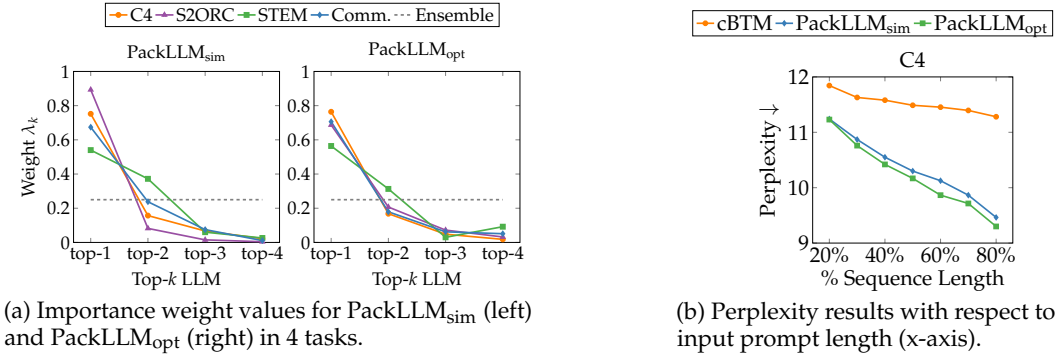
\begin{figure}[tb]
  \centering
    \begin{subfigure}[b]{0.5\textwidth}
        \resizebox{\linewidth}{!}{\definecolor{col1}{rgb}{0.60, 0.31, 0.64}
\definecolor{col2}{rgb}{0.30, 0.69, 0.29}
\definecolor{col3}{rgb}{0.22, 0.49, 0.72}
\definecolor{col4}{rgb}{0.89, 0.10, 0.11}
\definecolor{col5}{rgb}{1, 1, 0.8}

\begin{tikzpicture}
\tikzstyle{every node}=[font=\LARGE]

\begin{axis}[legend style={at={(.9,0.9),anchor=north east}},
             legend style={legend pos=outer north east,},  title={\packs},ylabel={Weight $\lambda_k$}, xlabel={Top-$k$ LLM}, every axis plot/.append style={ultra thick}, ymin=0, ymax=1, compat=1.5, xtick={1,2,3,4}, xticklabels={top-$1$,top-$2$,top-$3$,top-$4$},%
]

\addplot[mark=*,orange, error bars/.cd, y dir=both, y explicit] coordinates {
    (1, 0.752)
    (2, 0.157)
    (3, 0.067)
    (4, 0.022)
};

\addplot[mark=triangle*,col1, error bars/.cd, y dir=both, y explicit] coordinates {
    (1, 0.893)
    (2, 0.082)
    (3, 0.015)
    (4, 0.004)

};

\addplot[mark=square*,col2, error bars/.cd, y dir=both, y explicit] 
coordinates {
    (1, 0.540)
    (2, 0.372)
    (3, 0.060)
    (4, 0.026)
};

\addplot[mark=diamond*,col3, error bars/.cd, y dir=both, y explicit] 
coordinates {
    (1, 0.674)
    (2, 0.238)
    (3, 0.075)
    (4, 0.011)
};

\addplot[dashed, gray] coordinates {(1, 0.25)(4, 0.25)};

\end{axis}

\begin{axis}[legend columns=-1,legend style={at={(0.9,1.4),anchor=north east,}}, title={\packo},xlabel={Top-$k$ LLM}, every axis plot/.append style={ultra thick}, ymin=0, ymax=1, compat=1.5, xtick={1,2,3,4}, xticklabels={top-$1$,top-$2$,top-$3$,top-$4$},
             xshift=8cm%
]

\addplot[mark=*,orange, error bars/.cd, y dir=both, y explicit] coordinates {
    (1, 0.764)
    (2, 0.168)
    (3, 0.048)
    (4, 0.018)
};

\addlegendentry{C4 }

\addplot[mark=triangle*,col1, error bars/.cd, y dir=both, y explicit] coordinates {
    (1, 0.687)
    (2, 0.207)
    (3, 0.072)
    (4, 0.032)

};

\addlegendentry{S2ORC }

\addplot[mark=square*,col2, error bars/.cd, y dir=both, y explicit] 
coordinates {
    (1, 0.564)
    (2, 0.313)
    (3, 0.030)
    (4, 0.092)
};
\addlegendentry{STEM }

\addplot[mark=diamond*,col3, error bars/.cd, y dir=both, y explicit] 
coordinates {
    (1, 0.706)
    (2, 0.179)
    (3, 0.062)
    (4, 0.051)
};
\addlegendentry{Comm. }

\addplot[dashed, gray] coordinates {(1, 0.25)(4, 0.25)};
\addlegendentry{Ensemble}

\end{axis}

\end{tikzpicture}}
    \vspace{-0.2in}
    \caption{Importance weight values for \packs (left) and \packo (right) in 4 tasks.}
     \label{fig:lambda}
    \end{subfigure} \hfill
\begin{subfigure}[b]{0.36\textwidth}
    \resizebox{\linewidth}{!}{\definecolor{col1}{rgb}{0.60, 0.31, 0.64}
\definecolor{col2}{rgb}{0.30, 0.69, 0.29}
\definecolor{col3}{rgb}{0.22, 0.49, 0.72}
\definecolor{col3b}{rgb}{0.25, 0.55, 0.70}
\definecolor{col4}{rgb}{0.89, 0.10, 0.11}
\definecolor{col5}{rgb}{1, 1, 0.8}

\begin{tikzpicture}
\tikzstyle{every node}=[font=\huge]

\begin{axis}[legend columns=-1,legend style={at={(1.2,1.4),anchor=north east,}},  
 title={C4},ylabel={Perplexity $\downarrow$}, xlabel={\% Sequence Length}, every axis plot/.append style={ultra thick}, ymin=9, ymax=12, compat=1.5, xtick={20,  40,  60, 80}, xticklabels={20\%, 40\%,  60\%, 80\%},
]

\addplot[mark=*,orange, error bars/.cd, y dir=both, y explicit] coordinates {
    (20, 11.844)
    (30, 11.629)
    (40, 11.580)
    (50, 11.488)
    (60, 11.454)
    (70, 11.395)
    (80, 11.281)
};

\addlegendentry{cBTM }

\addplot[mark=diamond*,col3, error bars/.cd, y dir=both, y explicit] 
coordinates {
    (20, 11.24)
    (30, 10.87) 
    (40, 10.55)
    (50, 10.30)
    (60, 10.126)
    (70, 9.865)
    (80, 9.463)
};
\addlegendentry{\packs }

\addplot[mark=square*,col2, error bars/.cd, y dir=both, y explicit] 
coordinates {
    (20, 11.23)
    (30, 10.76) 
    (40, 10.42)
    (50, 10.168)
    (60, 9.865)
    (70, 9.715)
    (80, 9.300)
};
\addlegendentry{\packo }

\end{axis}

\end{tikzpicture}}
  \vspace{-0.2in}
    \caption{Perplexity results with respect to input prompt length (x-axis).}
     \label{fig:prompt}
    \end{subfigure}
  \caption{Analysis of \packs and \packo.}
\end{figure}
\Figref{fig:lambda} shows the derived importance weights $\lambda_k$ by the two approaches for four different tasks.
\packs calculates the weights based on perplexity scores (\Eqref{eq:packs}) and thus, fluctuations in perplexity scores among datasets and LLMs result in fluctuations in the induced weights. For example, the top-1 weights range within $[0.54, 0.89]$ and top-2 weights range within $[0.08, 0.37]$. On the other hand, \packo solves  an optimization of minimizing perplexity to determine weights  $\lambda_k$, which results in weights of smaller range. For example, the top-1 weights range within $[0.56, 0.76]$ and top-2 weights range within $[0.16, 0.31]$.

\Figref{fig:prompt} compares cBTM, \packs, and \packo with respect to the length of the input prompt. cBTM does not take into account how well the LLMs understand the input and shows minor improvements with longer prompts. On the other hand, both \packs and \packo significantly improve as we increase the length of the input. Moreover, \packo outperforms \packs due to its optimization framework, which specializes into understanding the input better. 

\begin{wraptable}{r}{0.5\textwidth}
\centering
\caption{Time Complexity Analysis.}
\vspace{-0.15in}
\label{tab:time}%
\resizebox{\linewidth}{!}{
    \begin{tabular}{l|cc}
        \toprule
       PPL Opt.  & Opt. Time $\downarrow$ (sec.) & Perplexity $\downarrow$ \\
       \#LLMs  & 3 / 5 & 3 / 5 \\
         \midrule
         \textbf{Grid Search} & \\
        None   &  \textcolor{teal}{0} /  \textcolor{teal}{0}  & \textcolor{purple}{11.96} /  \textcolor{purple}{11.54}  \\
        Greedy   &  \textcolor{teal}{2} /  \textcolor{teal}{4} &  \textcolor{teal}{11.77} /  \textcolor{teal}{11.40} \\
        Exhaustive   &  \textcolor{teal}{14} /  \textcolor{purple}{902}  &  \textcolor{teal}{11.76} /  \textcolor{teal}{11.38} \\
        \bottomrule
    \end{tabular}%

}
\end{wraptable}%

Table~\ref{tab:time} compares different optimization approaches for determining the weights $\lambda_k$ on the C4 dataset. 
\packs does not solve an optimization problem and thus does not result in any time overhead (row ``None'' in Table~\ref{tab:time}). However, it is not ensured that it can achieve the best performance (PPL). 
On the other hand, \packo employs a greedy solution to its optimization problem (greedy grid search), improving downstream performance (perplexity), while maintaining a linear time complexity with respect to the number of seed LLMs (\#LLMs), i.e., 2-4 secs when using 2-5 LLMs. Although exhaustive grid search can slightly improve performance compared to greedy grid search, its time cost can be prohibitive. For example, when using 5 LLMs, exhaustive grid search results in a explosive time cost of 902 secs.

\section{Conclusion}
We introduce \packllm, an effective method for test-time fusion that leverages each LLM’s expertise, given an input prompt. \packllm performs model fusion by
solving an optimization problem for determining each LLM’s importance, so that perplexity over the input prompt is minimized. We conduct experiments with over 100 total LLMs on a diverse set of
tasks. Experimental results show that (i) perplexity is a reliable measure for
LLM fusion, (ii) PackLLM outperforms test-time fusion baselines by 1.89\% accuracy points, and (iii) PackLLM can leverage new LLMs to improve performance over learning-based fusion approaches by 3.92--11.94\% accuracy points.

\bibliography{FusionLLM}
\bibliographystyle{colm2024_conference}
\newpage
\appendix
{\large \textbf{Appendix}}
\section{Greedy Algorithm}\label{app:alg}

We present the greedy algorithm for prompt perplexity minimization in Algorithm~\ref{alg:greedy}.

\begin{algorithm}[h]
\centering
\caption{Greedy Perplexity Minimization for Test-Time Fusion.} \label{alg:greedy}
\begin{algorithmic}[1]
   \STATE {\bfseries Input:} Test Prompt $P = (x_0, x_1, \dots, x_t)$, Seed LLMs $\{M_1, \dots, M_K\}$. \hfill\textcolor{brown}{\COMMENT{\textit{test-time}}}
   \item[]
   \FORALL{$k \in K$}
   \STATE Compute $\text{PPL}_k(P)$. \hfill\textcolor{brown}{\COMMENT{\textit{perplexity estimation}}}
   \ENDFOR
   \STATE Sort LLMs via: $[ M^\ast_1, \dots, M^\ast_K] = argsort \big( \text{PPL}_1(P), \dots, \text{PPL}_K(P) \big)$.

   \STATE Set $\lambda^\ast_1 = 1$ and $\lambda^\ast_{k>1} = 0$. \hfill\textcolor{brown}{\COMMENT{\textit{weight initialization}}}
   \item[]
   \STATE Compute logits $\vs^{(1)}$ using $M^\ast_1$. \hfill\textcolor{teal}{\COMMENT{\textit{forward pass}}}
   \FORALL{$k \in [1,K-1]$}
   \STATE Compute logits $\vs^{(k+1)}$ using $M^\ast_k$. \hfill\textcolor{teal}{\COMMENT{\textit{forward pass}}}
    
   \FOR{$\lambda \in [0,1, step=0.05]$ } 
   \STATE Compute $\gL_{\lambda} = -\frac{1}{t} \sum_i^t \log \softmax  \big( \lambda \vs^{(k)}(x_i | x_{<i}) + (1-\lambda) \vs^{(k+1)}(x_i | x_{<i}) \big)$. \hfill\textcolor{teal}{\COMMENT{\textit{grid search}}}
   \ENDFOR
   \STATE Select $\lambda^{(k)} = \argmin_\lambda \gL_{\lambda} $. \hfill\textcolor{teal}{\COMMENT{\textit{loss minimization}}}
    \STATE Set $\vs^{(k)} = \lambda^{(k)} \vs^{(k)} + (1-\lambda^{(k)}) \vs^{(k+1)}$. \hfill\textcolor{teal}{\COMMENT{\textit{greedy fusion}}}
    \STATE Set $\lambda^\ast_{<k+1} = \lambda^{(k)} \cdot \lambda^\ast_{<k+1}$ and  $\lambda^\ast_{k+1} = (1-\lambda^{(k)})$. \hfill\textcolor{teal}{\COMMENT{\textit{weight update}}}
    \ENDFOR
    
    \item[]
    \STATE {\bfseries Output:} Weights $\{\lambda^\ast_1, \dots, \lambda^\ast_K\}$ for test input $P$. \hfill\hfill\textcolor{olive}{\COMMENT{\textit{weights for $\{M^\ast_1, \dots, M^\ast_K\}$}}}
    
    \item[]
    \STATE {\bfseries Generation:} Sample with token probability $p(x_{t'} | x_{<t'}) = \softmax \big( s( x_{t'} | x_{<t'}) \big)$, where $ \vs = \lambda^\ast_1 \vs_{M^\ast_1} + \dots + \lambda^\ast_K \vs_{M^\ast_K}$. \hfill\textcolor{olive}{\COMMENT{\textit{LLM fusion}}}

\end{algorithmic}
\end{algorithm}

\section{Dataset Details} \label{app:data}

All datasets can be found in HuggingFace Hub. 

C4~\citep{roberts2019exploring} is s a publicly available
distribution of a Common Crawl snapshot. Following cBTM,  we use 168B tokens of the no blocklist version
(en.noblocklist) that is out of
distribution to the OPT model's pretraining corpus. S2ORC~\citep{lo2019s2orc} is a publicly available corpus of
full-text academic papers from the Semantic Scholar. The
corpus spans 20 fields of study (e.g., Biology, Computer
Science, Art). Following cBTM, the training data consists of 168B tokens over multiple epochs. For each dataset, we report language modeling perplexity on 200 randomly-sampled
held out documents. For each test document, the prompt P consists of the 32 first tokens or
the first 20\% of the sequence’s tokens, whichever is shorter; perplexity is evaluated on the
rest of tokens.

MMLU~\citep{hendrycks2021mmlu} consists of knowledge intensive tasks, such as STEM exams, that have a question and four multiple choice questions. The prompt template is ``\texttt{Question: \{QUESTION\} Choices: \{Choices\} Answer: }''. We use the default 5-shot examples and evaluate on the test data. 

We employ ARC~\citep{clark2018arc}, BoolQ~\citep{clark2019boolq}, HellaSwag~\citep{zellers2019hellaswag}, and OpenBookQA~\citep{mihaylov2018openbook} datasets to evaluate performance on commonsense reasoning. We use 0-shot prompts with the template ``\texttt{Question: \{QUESTION\} Choices: \{CHOICES\} Answer: }'' for ARC, BoolQ and OpenBookQA, and continuation ``\texttt{The topic is about \{TOPC\}. \{CTX-A\} \{CTX-B\}}'' for HellaSwag. We randomly sample 256 test data for evaluation. BoolQA is a binary classification, while ARC, HellaSwag, and OpenBookQA usually have four answer choices.

For PubMedQA~\citep{jin2019pubmedqa} and USMLE~\citep{jin2021usmle}, we follow \citep{cheng2023adaptllm}. The template is ``\texttt{Question: \{QUESTION\} Choices: \{CHOICES\} Answer: }'' for PubMedQA and ``\texttt{\{CONTEXT\} \{QUESTION\} \{ANSWER\}}'' for USMLE. These datasets have four-way multiple choice answers. We randomly sample 256 test data for evaluation and use 0-shot prompts. 

Regarding miscellaneous classification tasks, we experiment with AGNews~\citep{zhang2015character} (topic classification), SST2~\citep{socher2013recursive} (sentiment analysis), Amazon~\citep{mcauley2013hidden} (sentiment analysis), and SemEval~\citep{basile2019semeval} (Twitter hatespeech/sentiment). We follow ~\citep{mavromatis2023adaicl} which constructs a pool of 30 annotated data, and the 5 closest examples of the pool (via embedding similarity) are used as few shot examples for a test input. We evaluate on 256 randomly sampled test data. 

\section{Seed LLMs Details} \label{app:llms}
In this section, we provide the full list of the LLMs we use. 

\textbf{MMLU STEM}. We use the following LLMs:
\begin{itemize}
    \item 2 Seed LLMs: 
    \item \texttt{meta-llama/Llama-2-7b-hf} 
    \item \texttt{mistralai/Mistral-7B-v0.1} 
    \\
    \item 4 Seed LLMs:  
    \item \texttt{meta-llama/Llama-2-7b-hf}
    \item \texttt{mistralai/Mistral-7B-v0.1}
    \item \texttt{microsoft/phi-2} 
    \item \texttt{Deci/DeciLM-7B}
    \\
    \item 6 Seed LLMs: 
    \item \texttt{meta-llama/Llama-2-7b-hf}
    \item \texttt{mistralai/Mistral-7B-v0.1}
    \item \texttt{microsoft/phi-2}
    \item \texttt{Deci/DeciLM-7B}
    \item \texttt{BioMistral/BioMistral-7B}
    \item \texttt{EleutherAI/llemma-7b}
\end{itemize}

\textbf{Commonsense}. We use the following LLMs:
\begin{itemize}
    \item 2 Seed LLMs: Same as MMLU-STEM.
    \item 4 Seed LLMs: Same as MMLU-STEM.
    \item 6 Seed LLMs: 
    \item \texttt{meta-llama/Llama-2-7b-hf}
    \item \texttt{mistralai/Mistral-7B-v0.1}
    \item \texttt{microsoft/phi-2}
    \item \texttt{Deci/DeciLM-7B}
    \item \texttt{Wanfq/FuseLLM-7B}
    \item \texttt{openlm-research/open-llama-7b-v2}
\end{itemize}

\textbf{Medicine}. We use the following LLMs:
\begin{itemize}
    \item 3 Seed LLMs: 
    \item \texttt{AdaptLLM/medicine-LLM} (expert)
    \item \texttt{luodian/llama-7b-hf} (anti-expert)
    \item \texttt{meta-llama/Llama-2-13b-hf} (base)
    \\
    \item 4 Seed LLMs: 
    \item \texttt{BioMistral/BioMistral-7B}
    \item \texttt{llSourcell/medllama2-7b}
    \item \texttt{AdaptLLM/medicine-LLM}
    \item \texttt{chaoyi-wu/PMC-LLAMA-7B}
\end{itemize}

\textbf{Miscellaneous}.  We use 12 expert OPT-1.3B models on trained on the C4 dataset by~\citep{gururangan2023cbtm}. We combine together 8 cluster experts and 4 cluster experts. 

\textbf{FuseLLM LLMs}. FuseLLM uses the following 3 LLMs:
\begin{itemize}
    \item \texttt{openlm-research/open-llama-7b-v2}
    \item \texttt{mosaicml/mpt-7b}
    \item \texttt{meta-llama/Llama-2-7b-hf}
\end{itemize}

\textbf{FoE LLMs}. FuseLLM uses the following 15 LLMs:
\begin{itemize}
    \item \texttt{Aspik101/trurl-2-7b-pl-instruct\_unload}
    \item \texttt{Charlie911/vicuna-7b-v1.5-lora-mctaco}
    \item \texttt{Fredithefish/RedPajama-INCITE-Chat-3B-Instruction-Tuning-with-GPT-4}
    \item \texttt{GOAT-AI/GOAT-7B-Community}
    \item \texttt{TheTravellingEngineer/bloom-1b1-RLHF}
    \item \texttt{ashercn97/manatee-7b}
    \item \texttt{garage-bAInd/Platypus2-7B}
    \item \texttt{golaxy/gogpt-7b-bloom}
    \item \texttt{julianweng/Llama-2-7b-chat-orcah}
    \item \texttt{lmsys/vicuna-7b-v1.3}
    \item \texttt{lmsys/vicuna-7b-v1.5-16k}
    \item \texttt{medalpaca/medalpaca-7b}
    \item \texttt{rombodawg/LosslessMegaCoder-llama2-7b-mini}
    \item \texttt{togethercomputer/GPT-JT-6B-v0}
    \item \texttt{togethercomputer/GPT-JT-6B-v1}
\end{itemize}

\textbf{Recent LLMs}. As recently released LLMs, we use the following:
\begin{itemize}
    \item \texttt{meta-llama/Llama-2-7b-hf}
    \item \texttt{mistralai/Mistral-7B-v0.1}
    \item \texttt{microsoft/phi-2}
    \item \texttt{Deci/DeciLM-7B}
\end{itemize}

\textbf{Recent SLMs}. As the recently released SLMs, we use the following:
\begin{itemize}
    \item \texttt{google/gemma-2b}
    \item \texttt{microsoft/phi-2}
    \item \texttt{Qwen/Qwen1.5-1.8B}
\end{itemize}

\section{Full Results} \label{app:full-results}
Table~\ref{app:stem-results} provides full results on MMLU-STEM tasks. 

Table~\ref{app:common-results} provides full results on commonsense tasks.

Table~\ref{app:med-results} provided full results on medicine tasks. Further MMLU-Health results are provided in Table~\ref{tab:health}.

\begin{table*}
\centering
\caption{Full results on MMLU-STEM (College) tasks. }
\label{app:stem-results}%
\resizebox{0.6\columnwidth}{!}{
\begin{threeparttable}
    \begin{tabular}{l|ccccc|c}
        \toprule
        & Biology & Chemistry & CS & Math & Physics & \textbf{Average} \\
         \midrule
        & \multicolumn{5}{c}{2 Seed LLMs (14B)} \\

        Ensemble & 59.02 & 42.00 & 45.00 & 32.00 & 28.43 & 41.29\\
        Top1-PPL & 46.52 & 40.00 & 37.00 & 33.00 & 31.37 & 37.59 \\
        \textbf{\packs} & 53.47 & 45.00 & 45.00 & 35.00 & 32.35 & 42.16\\
        \textbf{\packo} & 54.17 & 46.00 & 45.00 & 33.00 & 33.33 & \textbf{42.30}\\
          \midrule
          
        & \multicolumn{5}{c}{4 Seed LLMs (23.7B)} \\
        
        Ensemble & 60.41 & 49.00 & 41.00 & 32.00 & 35.29 & 43.54\\
        Top1-PPL & 58.33 & 34.00 & 38.00 & 39.00 & 34.31 & 40.73 \\
        \textbf{\packs} & 61.80 & 41.00 & 47.00 & 37.00 & 34.31 & 44.22\\
        \textbf{\packo} & 63.88 & 40.00 & 41.00 & 41.00 & 39.21 & \textbf{45.02}\\
        \midrule 

        & \multicolumn{5}{c}{6 Seed LLMs (34.7B)} \\
        Ensemble & 58.33 & 43.00 & 44.00 & 34.00 & 38.23 & 43.51\\
        Top1-PPL & 58.33 & 39.00 & 39.00 & 37.00& 50.00 & 44.67 \\
        \textbf{\packs} & 63.19 & 42.00 & 44.00 & 35.00 & 38.23 & 44.48  \\
        \textbf{\packo} & 63.88 & 44.00 & 46.00 & 31.00 & 43.13 & \textbf{45.60}\\
        \midrule 
        & \multicolumn{5}{c}{3 Seed LLMs, same with FuseLLM (21B) } \\
        Ensemble & 40.97 & 33.00 & 34.00 & 26.00 & 23.52 & \textbf{31.49}\\
        Top1-PPL & 43.05 & 27.00 & 31.00 & 30.00 &  20.58 & 30.32\\
        \textbf{\packs} & 43.75 & 29.00 & 32.00 & 30.00 & 21.56 & 31.26 \\
        \textbf{\packo} & 44.44 & 31.00 &  32.00 & 31.00 & 17.64 & 31.21\\
        \midrule
        & \multicolumn{5}{c}{15 Seed LLMs, same with FoE (93B)} \\
        Ensemble & 46.06 & 34.75 & 36.77 & 33.40 & 26.47 & 35.49\\
        Top1-PPL & 47.91 & 39.00 & 36.00 & 30.00 & 24.50 & 35.48\\
        
        \textbf{\packs }& 53.47 & 37.00 & 37.00 & 41.00 & 25.49 & 38.79 \\
        \textbf{\packo} & 54.86 & 38.00 & 43.00 & 39.00 & 26.47 & \textbf{40.26}\\
        \midrule 
        & \multicolumn{5}{c}{3 Seed SLMs (6.5B)} \\
        Ensemble & 49.30 & 34.00 & 38.00 & 31.00 & 21.56 & 34.77\\
        Top1-PPL & 56.25 & 37.00 & 35.00 & 32.00 & 24.50 & 36.95 \\
        \textbf{\packs} & 55.55 & 35.00 & 37.00 & 31.00 & 26.47 & \textbf{37.00} \\
        \textbf{\packo} & 56.94 & 30.00 & 40.00 & 29.00 & 27.45 & 36.68 \\
        \midrule
        FuseLLM (7B) & 40.97 & 25.00  & 39.00 & 33.00 & 27.45 & 33.08\\
        FoE (93B) & 54.55 & 36.00 & 40.00 & 42.00 & 30.69 & 40.65\\

        \bottomrule
    \end{tabular}%
    \begin{tablenotes}
        \item 
    \end{tablenotes}

    \end{threeparttable}
}

\end{table*}%

\begin{table*}
\centering
\caption{Full results on Commonsense tasks. }
\label{app:common-results}%
\resizebox{0.6\columnwidth}{!}{
\begin{threeparttable}
    \begin{tabular}{l|cccc|c}
        \toprule
        & ARC-Challenge & BoolQ & HellaSwag & OpenBookQA&\textbf{Average} \\
         \midrule
        & \multicolumn{4}{c}{2 Seed LLMs (14B)} \\

        Ensemble &  54.29 & 80.46 & 74.60 & 55.85 & \textbf{66.30 }\\
        Top1-PPL & 41.01 & 78.51 &74.60  & 51.17 & 61.32\\
        \textbf{\packs} & 44.43 & 81.25 & 75.39 & 54.29 & 63.84\\
        \textbf{\packo} &  42.96 & 78.90 &  75.00 & 50.39 & 61.81\\
          \midrule

        & \multicolumn{4}{c}{4 Seed LLMs (23.7B)} \\
        Ensemble &  52.34  & 76.56 & 76.17 & 52.73 & 64.45\\
        Top-1 &   49.60 & 75.00 & 76.56 & 56.64 & 64.45\\
        \textbf{\packs} & 55.85 & 81.25 & 78.90 & 60.93 & \textbf{69.23}\\
        \textbf{\packo} &  51.95 & 78.51 & 77.73 & 62.50 & 67.67\\
          \midrule

        & \multicolumn{4}{c}{6 Seed LLMs (34.7B)} \\
        Ensemble  & 46.48 & 76.17 & 74.21 & 51.56 & 62.10 \\
        Top1-PPL & 48.04 & 80.01 & 76.56 & 56.64 & 65.31\\
        
        \textbf{\packs} & 53.31 & 83.20 & 78.90 & 	58.59 & \textbf{68.50 }\\
        \textbf{\packo} & 50.78 & 79.29 & 76.56 & 63.67 & 67.58\\
          \midrule
        & \multicolumn{4}{c}{3 Seed LLMs, same as FuseLLM (21B)} \\
        Ensemble  & 30.85 & 76.95 & 70.31 & 38.67 & 	54.19 \\
        Top1-PPL & 29.29 & 75.00 & 74.21 & 38.67 & 53.60\\
        \textbf{\packs} & 28.13 & 76.95 & 74.21 & 40.23 & \textbf{54.88}\\
        \textbf{\packo} &29.29 & 75.39 & 74.60 & 40.23 & 54.87\\
          \midrule

        & \multicolumn{4}{c}{15 Seed LLMs, same as FoE (93B)} \\
        Ensemble  & 47.90 &  73.39 & 72.60 & 52.06 & 61.48 \\
        Top1-PPL & 33.59 & 72.65 & 74.21 & 47.26 & 56.92\\
        \textbf{\packs} & 54.68 & 76.17 & 74.21 & 63.17 & \textbf{67.05} \\
        \textbf{\packo} & 48.82 & 69.53 & 75.39 & 55.07 & 62.20 \\

          \midrule
           & \multicolumn{4}{c}{3 Seed SLMs (6.5B)} \\
          Ensemble & 48.88 & 66.79 & 60.93 & 46.87 & 55.86\\
          Top1-PPL & 63.67 & 77.34 & 67.96 & 61.32 & 67.57\\ 
          \textbf{\packs} & 61.71 & 77.34 & 69.53 & 58.98 & 66.89 \\
          \textbf{\packo} & 62.50 & 79.68 & 69.14 & 61.71 & \textbf{68.25}\\
          \midrule

        FuseLLM (7B) & 36.32 &  77.73 & 75.78 &44.53 & 58.59\\
        
        \bottomrule
    \end{tabular}%
    \begin{tablenotes}
        \item 
    \end{tablenotes}

    \end{threeparttable}
}

\end{table*}%

\begin{table*}
\centering
\caption{Full results on Medicine tasks. }
\label{app:med-results}%
\resizebox{0,7\columnwidth}{!}{
\begin{threeparttable}
    \begin{tabular}{l|ccc|c}
        \toprule
         & MMLU-Health & PubMedQA & USMLE&\textbf{Average} \\
          & \textit{8 tasks, 5-shot} & \textit{0-shot} & \textit{0-shot} & (weighted) \\
         \midrule
        & \multicolumn{3}{c}{3 Seed LLMs (with anti-expert)} \\
        DExperts & 44.17 &66.40  &33.98 & 48.18 (45.37)\\

        Ensemble  & 40.31 & 68.75  & 31.03  & 46.69 (42.22)\\
        Top1-PPL  & 45.70 & 67.96 & 34.76 & 	49.47 (46.83)\\
        \textbf{\packs}   & 41.85 & 68.35 & 32.42 & 47.54 (43.55)\\
        \textbf{\packo}  & 46.13 & 69.14 & 33.98 & 49.75 (\textbf{47.21})\\
        \midrule
        & \multicolumn{3}{c}{4 Seed LLMs} \\

        Ensemble & 51.45 & 69.53 & 34.76 & 51.91 (51.59)  \\
        Top1-PPL & 51.61 & 54.29 & 33.98 & 46.62 (50.11) \\
        \textbf{\packs} & 53.22  & 67.57 & 33.98 & 51.59 (53.73)\\
        \textbf{\packo} & 54.65 & 67.96 & 33.59 & 52.06 (\textbf{53.88})\\

        \bottomrule
    \end{tabular}%
    \begin{tablenotes}
        \item 
    \end{tablenotes}

    \end{threeparttable}
}

\end{table*}%

\begin{table*}
\centering
\caption{Full results on MMLU-Health tasks. }
\label{tab:health}%
\resizebox{\columnwidth}{!}{
\begin{threeparttable}
    \begin{tabular}{l|cccccccc|c}
        \toprule
        & Anatomy & Clinical & CollegeMed & HumanAge & MedGenetics & Nutrition & ProMed & Virology & \textbf{Average} \\
        \midrule
        & \multicolumn{8}{c}{4 Seed LLMs (28B)} & \\

        Ensemble & 48.88 & 58.49 & 42.77 & 55.15 & 59.00 & 58.49 & 44.85 & 43.97 & 51.45\\
        Top1-PPL & 43.73 & 57.35 & 46.24 & 58.74 & 63.00 &  54.90 & 43.75 & 45.18  & 51.61\\
         \textbf{\packs} & 47.40 & 60.00 & 46.24 & 60.98 & 68.00 & 52.94 & 44.48 & 45.78 & 53.22\\
        \textbf{\packo} & 50.37 & 62.64 & 45.66 & 61.43 & 67.00 & 57.18 & 47.79 & 45.18 & \textbf{54.65}\\
        
        \bottomrule
    \end{tabular}%

    \end{threeparttable}
}

\end{table*}%

Table~\ref{app:misc-results} provides full results on miscellaneous classification tasks.

\begin{table*}
\centering
\caption{Results on Miscellaneous classification tasks. }
\label{app:misc-results}%
\resizebox{\columnwidth}{!}{
\begin{threeparttable}
    \begin{tabular}{l|cccccc|c}
        \toprule
        &  AGNews &  Ethos & TweetHate  & Amazon  & TweetComplain & SST2&\textbf{Average} \\
        & \textit{Topic}  & 

        \textit{Hatespeech} & \textit{Hatespeech} & 
        \textit{Sentiment} & 
        \textit{Sentiment} & 
        \textit{Sentiment} & \\
         \midrule
        & \multicolumn{7}{c}{12 Seed LLMs } \\
        cBTM & 58.98  & 50.39 & 46.09 & 70.70 &  78.00 &67.57 & 61.95 \\
        Ensemble & 62.10 & 51.17 & 46.48 & 69.92 & 76.00&69.53 & 62.53\\
        Top1-PPL & 62.89 &  51.17 & 45.31  & 74.21 & 78.00 &72.17 & 63.95\\
        \textbf{\packs} & 61.71&  51.56 & 46.48 & 69.92 & 80.00 &69.53 & 63.20\\
        \textbf{\packo} & 60.93 &  55.07 & 46.09 & 70.31 & 78.00& 69.92 & 63.38\\

        \bottomrule
    \end{tabular}%
    \begin{tablenotes}
        \item 
    \end{tablenotes}

    \end{threeparttable}
}

\end{table*}%

\end{document}